\documentclass{article}
\usepackage{nips13submit_e,times}
\usepackage{color}
\usepackage{graphicx,caption,subcaption,paralist}
\usepackage{amsmath,amssymb} 
\usepackage{xspace}

\usepackage{microtype}

\usepackage[unicode=true,pdfusetitle,
 bookmarks=true,bookmarksnumbered=false,bookmarksopen=false,
 breaklinks=false,pdfborder={0 0 1},backref=section,colorlinks=true]
 {hyperref}

\makeatletter
\DeclareRobustCommand\onedot{\futurelet\@let@token\@onedot}
\def\@onedot{\ifx\@let@token.\else.\null\fi\xspace}

\def\ie{\emph{i.e}\onedot} 
 
\def\etc{\emph{etc}\onedot}  
 
\def\etal{\emph{et al}\onedot}
\makeatother

\iftrue
\newcommand{\todo}[1]{{\textcolor{red}{[[TODO: #1]]}}}
\newcommand{\outline}[1]{{\textcolor{red}{[[#1]]}}}
\newcommand{\commenttext}[1]{\textcolor{blue}{[[#1]]}}
\newcommand{\commentfoot}[1]{\footnote{\textcolor{red}{#1}}}
\newcommand{\commentselfoot}[2]{{\textcolor{blue}{#1}}\comment{#2}}
\newcommand{\commentselrep}[2] {{\textcolor{blue}{#1}} {\textcolor{green}{[[\textit{#2}]]}}}
\else
\newcommand{\todo}[1]{}
\newcommand{\outline}[1]{}
\newcommand{\commenttext}[1]{}
\newcommand{\commentfoot}[1]{}
\newcommand{\commentselfoot}[2]{}
\newcommand{\commentselrep}[2]{}
\fi

\newcommand{\bdY}{\mathbf{Y}}

\title{Deep and Wide Multiscale Recursive Networks\\for Robust Image Labeling}

\author{
Gary B. Huang and Viren Jain \\
Janelia Farm Research Campus \\ 
Howard Hughes Medical Institute \\
19700 Helix Drive, Ashburn, VA, USA \\
\texttt{\{huangg, jainv\}@janelia.hhmi.org} 
}

\nipsfinalcopy 

\begin{document}
\maketitle

\begin{abstract}

Feedforward multilayer networks trained by supervised learning have recently demonstrated state of the art performance on image labeling problems such as boundary prediction and scene parsing. As even very low error rates can limit practical usage of such systems, methods that perform closer to human accuracy  remain desirable. In this work, we propose a new type of network with the following properties that address what we hypothesize to be limiting aspects of existing methods: (1) a `wide' structure with thousands of features, (2) a large field of view, (3) recursive iterations that exploit statistical dependencies in label space, and (4) a parallelizable architecture that can be trained in a fraction of the time compared to benchmark multilayer convolutional networks. For the specific image labeling problem of boundary prediction, we also introduce a novel example weighting algorithm that improves segmentation accuracy. Experiments in the challenging domain of connectomic reconstruction of neural circuity from 3d electron microscopy data show that these ``Deep And Wide Multiscale Recursive'' (DAWMR) networks lead to new levels of image labeling performance. The highest performing architecture has twelve layers, interwoven supervised and unsupervised stages, and uses an input field of view of 157,464 voxels ($54^3$) to make a prediction at each image location.  We present an associated open source software package that enables the simple and flexible creation of DAWMR networks. 
\end{abstract}

\section{Introduction}

Image labeling tasks generate a pixel-wise field of predictions across an image space. In boundary prediction, for example, the goal is to predict whether each pixel in an image belongs to the interior or boundary of an object~\cite{Martin:2004}; in scene parsing, the goal is to associate with each pixel a multidimensional vector that denotes the category of object to which that pixel belongs~\cite{farabet2012scene}. These types of tasks are distinguished from traditional object recognition, for which pixel-wise assigments are usually irrelevant and the goal is to produce a single global prediction about object identity.

Densely-labeled pixel-wise ground truth data sets have recently been
generated for image labeling tasks that were traditionally solved by
entirely hand-designed methods~\cite{Martin:2004}.  This has enabled
the use of learning methods that require extensive supervised
parameter learning. As a result, a common class of methods, supervised
multilayer neural networks, have recently been found to excel at image
labeling and object recognition tasks. This approach has led to
state-of-the-art and in some cases breakthrough performance on a
diverse array of problems and data
sets~\cite{farabet2012scene,krizhevsky2012imagenet,Jain:2010,ciresan2012deep,jain2008natural}.
Despite these improvements, for most practical applications even
higher accuracy is required to achieve reliable automated image
analysis. For example, in the main application studied in this paper,
reconstruction of neurons from nanometer-resolution electron
microscopy images of brain tissue, even small pixel-wise error rates
can catastrophically deteriorate the utility of automated
analysis~\cite{jain2010machines}.

In this paper, we identify limitations in existing multilayer network architectures, propose a novel architecture that addresses these limitations, and then conduct detailed experiments in the domain of connectomic reconstruction of electron microscopy data. The primary contributions of our work are: 
\begin{compactenum}
	\item A `wide' and multiscale core architecture whose labeling accuracy exceeds a standard benchmark of a feedforward multilayer convolutional network. By exploiting parallel computing on both CPU clusters and GPUs, the core architecture can be trained in a day, compared with two weeks for a GPU implementation of the convolutional network.
	\item A recursive pipeline consisting of repeated iterations of the core architecture. Through this recursion, the network is able to increase the field of view used to make a prediction at a given image location and exploit statistical structure in label space, resulting in substantial gains in accuracy.
	\item A computationally efficient scheme for weighting training set examples in the specific image labeling problem of boundary prediction. This approach, which we refer to as `local error density' (LED) weighting, is used to focus supervised learning on difficult and topologically relevant image locations, and leads to more useful boundary predictions results.
\end{compactenum}

\section{Networks for Image Labeling: Prior Work and Desiderata}

Multilayer networks for visual processing combine filtering,
normalization, pooling, and subsampling operations to extract features
from image data. Feature extraction is followed by additional
processing layers that perform linear or nonlinear classification to
generate the desired prediction variables
\cite{Jarrett2009best}. Farabet \etal recently adapted
convolutional networks to natural image scene labeling by training 2d
networks that process the image at multiple scales
~\cite{farabet2012scene}. Boundary prediction in large-scale
biological datasets has been investigated using 3d architectures that
have five to six layers of
processing~\cite{Jain:2007,Turaga:2010uq,Jain:2010,Turaga:2009} and,
in the work of Ciresan \etal, 2d architectures with pooling operations
and ensembles of multiple networks~\cite{ciresan2012deep}. These
studies have shown that multilayer networks often outperform
alternative machine learning methods, such as MRFs and random forest
classifiers. We hypothesize that image labeling accuracy could be
further improved by a network architecture that simultaneously
addresses all of the following issues:

{\bf Narrow vs wide feature representations:} The number of features
in each layer of a network plays a major role in determining the
overall capacity available to represent different aspects of the input
space. Most multilayer models for image labeling have thus far been
relatively `narrow', \ie, containing a small number of features in
each layer. For example, networks described in Jain
\etal and Farabet \etal used
respectively 12 and 16 features in the first layer, while those in
Ciresan \etal used 48. We would like to
transition to much wider architectures that utilize thousands of
features. 

{\bf Large field of view:} Local ambiguity in an image interpretation
task can be caused by noise, clutter, or intrinsic uncertainty
regarding the interpretation of some local structures. Global image
information can be used to resolve local ambiguity, and thus effective
integration of image data over large fields of view is critical to
solving an image labeling task.  In multilayer visual processing
architectures, there are a variety of factors that determine the
effective size of the field of view used to compute a prediction for a
specific pixel location: filter size, network depth, pooling
structure, and multiscale processing pathways. Experiments in this
work and others suggest that appropriate usage of all of these
architectural components is likely to be necessary to achieve highly
accurate image labeling.

While the 2d architecture proposed for scene labeling in Farabet
\etal is already multiscale, converting the
architecture to utilize 3d filters lengthens training time into weeks
or more. The 3d boundary prediction networks in Jain
\etal and Turaga \etal have also
been augmented with multiscale capabilities, but these modifications
lengthen training times from weeks into months.  

{\bf Modeling and exploiting statistical structure in labels:} In the
multilayer networks introduced thus far, predictions about neighboring
image locations are nearly independent and become potentially
correlated only due to a dependence on overlapping parts of the input
image. However, in image labeling tasks there is usually a substantial
amount of statistical structure among labels at neighboring image
locations. This observation suggests that image labeling is a
\emph{structured} prediction problem in which statistics among output
variables should be explicitly modeled \cite{tsochantaridis2005large}.
Markov random field (MRF) image models are an example of a generative approach to structured prediction. These methods consist of an observation model $p(X|\bdY)$, encoding the conditional distribution of the image $X$ given the labels $\bdY$, and a prior model $p(\bdY)$, specifying the distribution over different label configurations. Given a noisy input image $X$, inference for $p(\bdY|X)$ can thus involve both image-dependent aspects (to invert the observation model) as well as interactions among the random variables $Y \in \bdY$ that reflect prior statistics on valid label configurations~\cite{li1994markov}.

Multilayer network models for image analysis typically outperform
MRFs, as the computational expense associated with probabilistic
learning and inference substantially restricts the modeling capability
of MRF models that are practical to work
with~\cite{Jain:2007,jain2008natural}. An alternative approach is
pursued by Farabet \etal, in which a multilayer network is augmented
by a simple three-parameter CRF post-processing step designed to
`clean up' classifier predictions. In this
work, we propose and investigate a recursive approach in which outputs
from one network become input for a subsequent network, thereby
allowing for explicit and powerful modeling of statistics among output
predictions.

{\bf Reasonable training time:} We regard it as critical that a
network can be learned in a reasonable amount of wall-clock time
(within a few days at most, but more ideally within hours). Many
existing approaches could conceptually be scaled up to address the
limitations that we discuss, but would then require weeks or more in
order to train. Such long training times can prohibit certain usage
scenarios (for example, interactively adding new labeled data based on
rapid classifier retraining~\cite{sommer2011ilastik}). More generally,
experimenting in the space of different cost functions, architectures,
labeling strategies, \etc, is only feasible if a single experiment can
be performed in a reasonable duration of time.  To achieve a
reasonable training time, in this paper we assume access to both
graphics processing units (GPUs) and multi-core CPUs or cluster
computing environments. Many recent and notable results in machine
learning would not have been possible without parallel computing on
GPUs or
CPUs~\cite{krizhevsky2012imagenet,coates2012emergence,pinto2009high}.

\section{Deep and Wide Multiscale Recursive Networks} \label{sec:dawmr_nets}

We formalize the image labeling problem as follows: given an input
image $I$, we want to predict at each location $l \in I$ a vector of
labels $\bdY_l$. For the main data set considered, $I$ is a three
dimensional volume of size on the order of $1000^3$, and with each
location is associated a vector of 3 labels ($|\bdY_l|=3)$, indicating
3d neighbor connectivity (see
Section~\ref{sec:training_test_evaluation} for details of the data and
labels). For ease of notation we will treat the location $l$ as a
single dimension with regard to operations discussed later, such as
pooling, but in practice such operations are applied in 3d.

\begin{figure}[t]
	\centering 
	\includegraphics[width=4.75in]{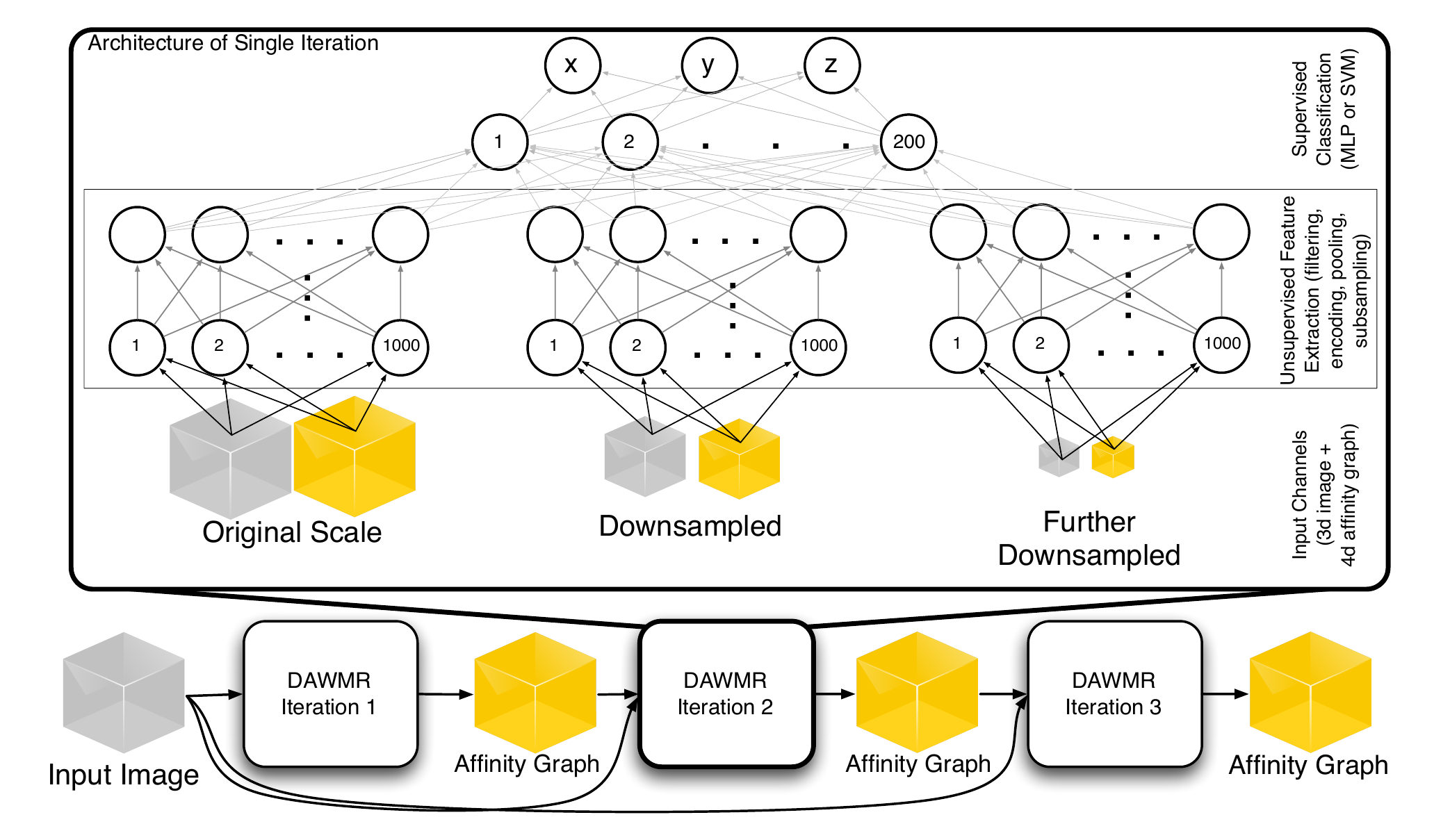} \caption{\small Illustration of a DAWMR network with three recursive iterations.} \label{fig:dawmr_illustration} 
\end{figure}

In this section, we describe our proposed method for image labeling,
Deep and Wide Multiscale Recursive (DAWMR) networks. DAWMR networks
process images by recursive iteration of a core network
architecture. Overall, a DAWMR network may have dozens of individual
processing layers between the raw input image and final labeling
output. A schematic overview of a typical DAWMR network is given in
Figure~\ref{fig:dawmr_illustration}.

\subsection{Single Iteration Core Architecture} \label{sub:core-network-architecture}

The core network architecture in each iteration consists of two
sequential processing modules: feature extraction and
classification. These stages are conceptually distinct, learned using
differing levels of supervision, and implemented using different
parallel computing strategies.


{\bf Feature Extraction:} Given a location $l$, the feature extraction
module produces $h_l$, a representation of the input image centered at
$l$. These representations are subsequently passed to the classifier
to learn the specific image labeling that is encoded in the training
data. (Generally $h_l$ is normalized such that each feature has zero
mean and unit standard deviation prior to being passed to the
classifier.)

At a high level, each feature extraction module consists of multiple
processing layers of feature encoding using vector quantization (VQ),
with intermediate layers that apply operations such as pooling,
subsampling, and whitening. An entire set of processing layers can be
replicated within a single module to process the image at multiple
different downsampled scales (where downsampling is achieved by simple
averaging).
We take advantage of the recent observation that unsupervised
clustering and dictionary learning techniques can be used to
efficiently learn thousands of features for image recognition
tasks~\cite{coates2011analysis,coates2011importance}. Following Coates
and Ng~\cite{coates2011importance}, the core vector quantization
component in the feature extraction module consists of a dictionary
learned using a greedy variant of orthogonal matching pursuit (OMP-1)
and encoding using soft-thresholding with reverse polarity.

Given a learned dictionary for performing vector quantization, we can
produce an encoding $f_i$ centered at a location $i$. We consider two
contrasting methods for forming a final hidden representation $h_l$
from these encoding $f_i$.  The first method uses the encoding itself
as the representation, at various pixel locations centered at $i$. In
what we call an $m$ receptive field (RF) architecture, the hidden
representation $h_l$ is formed by concatenating $m$ features, $h_l =
\{f_{l-\frac{m}{2}}, \ldots, f_{l+\frac{m}{2}}\}$. This is similar to
the convolutional network architectures used
in~\cite{Jain:2007,jain2011learning,Turaga:2010uq}; in those networks,
classification is based on input from all feature maps in the final
hidden layer from feature map values within a $5^3$ pixel window
centered at the location being classified.
The second method is a foveated representation that incorporates
pooling operations. Given some neighborhood size $m$, we first perform
max pooling over the neighborhood, $(g_l)_j =
\max_{i=l-\frac{m}{2}}^{l+\frac{m}{2}} (h_i)_j$. The foveated
representation is then the concatenation of the feature encoding
centered at $l$ and the pooled feature, $h_l = \{ f_l, g_l\}$. We also
experimented with average pooling but found max pooling to give better
results in general.

We note that an $m$ RF architecture and a foveated representation with
a pooling neighborhood of size $m$ have the same field of view of the
data. However, if the dimensionality of the encoding is $|f_i| = k$,
then the dimensionality of the hidden representation using an $m$ RF
architecture is $|h_l| = mk$, whereas with a foveated representation
$|h_l| = 2k$. Therefore, these two methods lie at opposite ends of the
spectrum of `narrow' versus `wide' network architectures. Given a
fixed hidden representation dimensionality $|h_l| = d$, the $m^3$ RF
architecture will have a narrow VQ dictionary ($\frac{d}{m^3}$)
whereas the foveated representation will be able to support a wider VQ
dictionary ($\frac{d}{2}$).

{\bf Classification:} Following the feature extraction module, we have
a standard supervised learning problem consisting of features $h_l$
and labels $\bdY_l$. For the classification module, we use a
multilayer perceptron (MLP) with a single hidden layer trained by
mini-batch stochastic gradient descent~\cite{hinton2012improving}. In
preliminary experiments, we found that an MLP outperformed a linear
SVM. For image labeling problems that involved predicting multiple
labels at each location $l$, we also found that using a single MLP
with multiple output units outperformed an architecture with multiple
single output MLPs.


{\bf Recursive Application of Core Network:} In recursive approaches
to prediction, a classifier is repeatedly applied to an input (and
previous classifier results) to build a structured interpretation of
some data. Pioneering work established graph transformer networks for
solving segmentation and recognition tasks arising in automated
document processing~\cite{bottou1997global}. More recently, recursive
approaches have been revived for superpixel
agglomeration~\cite{socher2011parsing,jain2011learning} and text
parsing~\cite{collobert2011deep}.

In image labeling tasks, each pixel in an input image generates a scalar or vector output that encodes predictions about the variables of interest. A straightforward way to directly model statistics\emph{ similar} to the labels is to use the output of an initial iteration of the architecture described in Section \ref{sub:core-network-architecture} and provide that ``network iteration 1'' ($N_{1}$) output as input to another instance of such an architecture ($N_{2}$). The $N_{1}$ output is accompanied by the raw image, and thus feature extraction and subsequent predictions from $N_{2}$ are based upon structure in both the original image as well as the output representation from $N_{1}$. Recursive construction of such classifiers is repeated for $N_{3},...,N_{k}$, where $k$ is as large as computation time permits or cross-validation performance justifies. 

Each additional recursive iteration also increases the overall field of view used to predict the output at a particular pixel location. Thus, recursive processing enables DAWMR networks to simultaneously model statistical regularities in label-space and use increasing amounts of image context, with the overall goal of refining image labeling predictions from one iteration to the next. 

\section{Boundary Prediction Experiments} \label{sec:experiments}

We performed detailed experiments in the domain of boundary prediction
in electron microscopy images of neural tissue. This application has
significant implications for the feasability of `connectomics', an
emerging endeavour in neurobiology to measure large-scale maps of
neural circuitry at the resolution of single-synapse
connectivity~\cite{lichtman2011big}. 
Reconstruction is currently the bottleneck in large-scale mapping
projects due to the slow rate of purely manual reconstruction
techniques~\cite{jain2010machines}. Fully-automated methods for
reconstruction would therefore be ideal. Current pipelines typically
begin with a boundary prediction step, followed by oversegmentation of
the resulting boundary map, and finally application of an
agglomeration algorithm to piece together object
fragments~\cite{andres20123d,jain2011learning}. Improvements in
boundary prediction are desirable, as certain types of errors (such as
subtle undersegmentation) can sometimes be difficult to correct during
later steps of the reconstruction pipeline. 

\subsection{Experimental Setup}

Here we describe the details of the the image data and training/test
sets.  Experiments were run using our parallel computing software
package, available
online\footnote{\url{http://sites.google.com/site/dawmrlib/}}; for
more details see Section~\ref{sec:implementation}.

\subsubsection{Image Acquisition}

Neuropil from \emph{drosophila melanogaster} was imaged using focused ion-beam scanning electron microscopy (FIB-SEM \cite{knott2008serial}) at a resolution of $8x8x8$ nm. The tissue was prepared using high-pressure freeze substitution and stained with heavy metals for contrast during electron microscopy. As compared to traditional electron microscopy methods such as serial-section transmission electron microscopy (ssTEM), FIB-SEM provides the ability to image tissue at very high resolution in all three spatial dimensions. Isotropic resolution at the sub-$10$nm scale is particularly advantageous in drosophila due to the small neurite size that is typical throughout the neuropil.

\subsubsection{Training, Test Sets} \label{sec:training_test_evaluation}

Two image volumes were obtained using the above acquisition
process. The first volume was used for training and the second for
both validation and testing. Initially, human annotators densely
labeled subvolumes from both images. These labels form a preliminary
training set, referred to in the sequel as the small training set (5.2
million labels), and the validation set (16 million labels),
respectively. Afterward, an interactive procedure was used wherein
human annotators `proofread' machine-generated segmentations by
visually examining dense reconstructions within small subvolumes and
correcting any mistakes. The proofreading interface enabled annotators
to make pixel-level modifications. The proofread annotations were then
added to the small training set and validation set to form the `full'
training set (120 million labels) and test set (46 million labels),
respectively.\footnote{The validation set is a subset of the test
  set. Measuring segmentation accuracy requires large densely-labeled
  subvolumes, and due to the expense in obtaining such data, we
  believe that measuring final test results on a larger set of data is
  more valuable for evaluation than splitting off a subset to only be
  used as validation.} 
%

\begin{figure}
  \centering
  \begin{subfigure}{0.17\textwidth}
    \includegraphics[width=\textwidth]{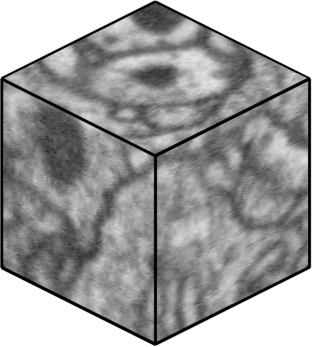}
    \caption*{image data\\~}
  \end{subfigure}
  \begin{subfigure}{0.13\textwidth}
    \centering
    \frame{\includegraphics[width=\textwidth]{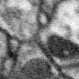}}
    \caption*{image data\\$z$-slice\\~}
  \end{subfigure}
  \begin{subfigure}{0.13\textwidth}
    \frame{\includegraphics[width=\textwidth]{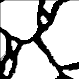}}
    \caption*{ground-truth affinity graph\\$z$-slice}
  \end{subfigure}
  \begin{subfigure}{0.13\textwidth}
    \centering
    \frame{\includegraphics[width=\textwidth]{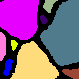}}
    \caption*{ground-truth segmentation\\$z$-slice}
  \end{subfigure}
  \begin{subfigure}{0.17\textwidth}
    \includegraphics[width=\textwidth]{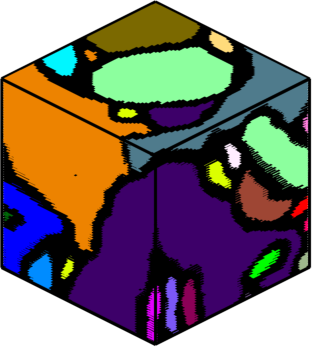}
    \caption*{ground-truth segmentation}
  \end{subfigure}
  \caption{\small A $79^3$ subcube of the test data and $z$-slice taken from
    the center of the cube.  Shown are the original image data,
    ground-truth affinity graph, and ground-truth segmentation.}
  \label{fig:data_examples}
\end{figure}

The labels are binary and indicate connectivity of adjacent voxels,
where positive labels indicate that two voxels belong to the same
foreground object and negative labels indicate that two voxels belong
to differing objects or both belong to the background. As the image
volume is three dimensional, each location is associated with a vector
of 3 labels in each direction. The set of such labels (or their
inferred probabilistic values) over an image volume is referred to as
the affinity graph. By specifying connectivity through an affinity
graph it is possible to represent situations (such as directly
adjacent, distinct objects) that would be impossible to represent with
a more typical pixel-wise exterior/interior labeling
\cite{Turaga:2009}.  Figure~\ref{fig:data_examples} shows a 2d slice
of the test image data, affinity graph, and segmentation.

\setlength{\fboxsep}{0pt}
\setlength{\fboxrule}{1pt}

\subsubsection{Evaluation Measures}

Given a densely-labeled ground truth segmentation, one can measure performance in two ways: classification metrics on binary representations of the segmentation (such as a boundary map or affinity graph), or segmentation-based measures that interpret the volume as a clustering of pixels. In this work, we report both types of measures. 

Boundary prediction performance is reported by treating affinity graph edge labeling as a standard binary classification task, where we compute results for each edge direction separately and then average the results over the three edge directions. As the ground truth has a class imbalance skewed toward positive (connected) edges, we report balanced class accuracy (\emph{bal-acc}: $0.5 \cdot \text{accuracy on positive edges} + 0.5 \cdot \text{accuracy on negative edges}$). We also compute area under the receiver operating characteristic  curve, when varying the decision threshold for classifying positive/negative edges (\emph{AUC-edge}).

One can also segment a ground truth affinity graph into clusters of connected voxels, forming a set of foreground objects and background. By segmenting an inferred affinity graph (whose labels may be real-valued) at a particular threshold, one can follow the same procedure to form an inferred clustering. We supplement the connected components segmentation by `growing out' segmented objects until they touch each other, using a marker-based watershed algorithm adapted to affinity graphs. Segmentation performance is then measured by computing the Rand Index \cite{Unnikrishnan:2007}. We report an area under the curve measure that integrates performance over different binarization thresholds (\emph{AUC-RI}), as well as a maximum score (\emph{max RI}).

\subsection{Model Selection Experiments on a Validation Set}

Like other deep, multilayer architectures, DAWMR networks have a
number of model/architecture parameters that can be varied. In this
section, we perform model selection experiments with the validation
set. Unless explicitly stated otherwise, our experiments use the
following set-up: the feature extraction modules produce a feature
representation of dimension $h^u_l = 8000$, individual filters use 3d
$5^3$ patches, and classification is performed using an MLP with a
single hidden layer with 200 hidden units and trained with a balanced
sampling of positive and negative training examples.

\subsubsection{Single-Iteration Classifiers and Comparison With Convolutional Networks} \label{sec:single_iter_classifiers}

We begin by evaluating performance of single-iteration DAWMR classifiers and a supervised convolutional network. We consider five DAWMR architectures: $5^3$ RF, single-scale vector quantization without pooling (SS), single-scale VQ with foveated representation (SS-FV), and multiscale VQ with foveated representation (MS-FV). We also test a version of the SS-FV architecture with 2d filters (other architectures use 3d filters). For both architectures using a foveated representation, we pool over a $5^3$ neighborhood, and thus the $5^3$ RF and SS-FV architectures have the same field of view. Table~\ref{tbl:single_iter_unsup_arch} provides an overview of the architectures.
\begin{table}
	[htbp] \centering 
	\begin{tabular}
		{|l|ccccc|} \hline arch. & $5^3$ RF & SS & SS-FV-2d & SS-FV & MS-FV \\
		\hline VQ dict. size & 32 & 4000 & 2000 & 2000 & 1000 \\
		field of view & $9^3$ & $5^3$ & $9^2$ & $9^3$ & $18^3$ \\
		feature dims & 8000 & 8000 & 8000 & 8000 & 8000 \\
		\hline 
	\end{tabular}
	\caption{\small Vector quantization dictionary size (\ie number of feature maps) and field of view for feature extraction modules in model selection experiments. The multiscale MS-FV architecture uses two dictionaries of size 1000, one for each scale. Table \ref{tbl:model_comparison} provides details regarding number of free parameters in the network architectures.} \label{tbl:single_iter_unsup_arch} 
\end{table}

Validation performance of single iteration DAWMR networks using the above feature extraction architectures is shown in Table~\ref{tbl:single_iter_classifiers_cn}. The performance of a supervised convolutional network (CN) is also provided, and represents a strong baseline for comparison; identical types of networks have been extensively used in recent studies involving boundary prediction in 3d electron microscopy data \cite{Jain:2010, Turaga:2010uq, helmstaedter2013connectomic}. The CN used in our experiment has 5 hidden layers, 16 feature maps per layer, all-to-all connectivity between hidden layers, and a filter size of $5^3$. The CN was trained on a GPU with an implementation based on the CNS framework \cite{Mutch:2010}. 

\begin{table}
	[htbp] \centering 
	\begin{tabular}
		{|l|ccc|cccc|} \hline & \multicolumn{3}{c|}{training set (small: 5M)} & \multicolumn{4}{c|}{validation set: 16M examples} \\
		network & bal-acc & AUC-edge & tr. time & bal-acc & AUC-edge & AUC-RI & max RI \\
		\hline CN & 0.9555 & 0.9872 & 2 weeks & 0.8322 & 0.8873 & 0.6692 & 0.8549 \\
		\hline $5^3$ RF & 0.8976 & 0.9628 & 1.5 days & 0.8200 & 0.8933 & 0.6764 & 0.9194 \\
		SS & 0.8565 & 0.9386 & 1.5 days & 0.7946 & 0.8859 & 0.6569 & 0.8757 \\
		SS-FV-2d & 0.8844 & 0.9583 & 1.5 days & 0.8024 & 0.8888 & 0.6537 & 0.8496 \\
		SS-FV & 0.9223 & 0.9799 & 1.5 days & 0.8129 & 0.8981 & 0.6799 & 0.9049 \\
		MS-FV & 0.9623 & 0.9935 & 1.5 days & 0.8305 & 0.9011 & 0.6796 & 0.9086 \\
		\hline MS-FV-DO & 0.9497 & 0.9899 & 1.5 days & \textbf{0.8372} & \textbf{0.9196} & \textbf{0.7119} & \textbf{0.9327} \\
		\hline 
	\end{tabular}
	\caption{\small Validation performance of single iteration (non-recursive) DAWMR classifiers for various feature extraction architectures, and comparison with a multilayer convolutional network (CN). SS: single-scale, MS: multiscale, FV: foveated, DO: drop-out. All architectures use 3d filters except SS-FV-2d.} \label{tbl:single_iter_classifiers_cn} 
\end{table}

The multiscale foveated architecture (MS-FV) achieves slightly better results than the convolutional network on most metrics, for both the training and test set. The DAWMR classifier is also learned in an order of magnitude less time than the convolutional network. Adding drop-out regularization (MS-FV-DO) improves performance of the single iteration DAWMR classifier even further~\cite{hinton2012improving}. 

\subsubsection{Recursive Multiscale Foveated Dropout (MS-FV-DO) Architecture} \label{sec:recursive_classifiers} 

A specific core architecture can be recursively applied over multiple
iterations, as discussed at the end of
Section~\ref{sub:core-network-architecture}.
In this section we experiment with this approach using the multiscale
foveated dropout (MS-FV-DO) architecture.  For the second and third
iteration classifiers, which accept as input both an affinity graph as
well as the original image, there is a model selection choice related
to whether filters in the feature extraction stage receive input from
only the image, only the affinity graph, or both. We found that
dividing the set of features into an equal number which look
exclusively at each type of input channel worked better than having
all filters receive input from all input channels.

\begin{table}
	[htbp] \centering 
	\begin{tabular}
		{|cc|cc|cccc|} \hline & & \multicolumn{2}{c|}{training set (large)} & \multicolumn{4}{c|}{validation set: 16M examples} \\
		model & iter & bal-acc & AUC-edge & bal-acc & AUC-edge & AUC-RI & max RI \\
		\hline MS-FV-DO & 1 & 0.9292 & 0.9809 & 0.8833 & 0.9497 & 0.7171 & 0.9565 \\
		MS-FV-DO & 2 & 0.9447 & 0.9858 & 0.8975 & 0.9551 & 0.7276 & \textbf{0.9691} \\
		MS-FV-DO & 3 & 0.9473 & 0.9883 & \textbf{0.9024} & \textbf{0.9628} & \textbf{0.7356} & \textbf{0.9691} \\
		\hline 
	\end{tabular}
	\caption{\small Performance of multiscale foveated architecture over multiple recursive iterations.} \label{tbl:recursive_noweighting} 
\end{table}

Table \ref{tbl:recursive_noweighting} shows the results from recursive
application of the MS-FV-DO architecture. Note that each iteration
learns its own unsupervised and supervised parameters, thereby
tripling the model parameters used to generate the final output, and
that each iteration adds $18^3$ to the total field of view used to
generate an output prediction by the third iteration classifier
($54^3$).  Recursive experiments were limited to three
iterations.\footnote{Additional iterations would require a field of
  view so large that significant amounts of labeled data in the
  training and validation set would no longer be usable due to
  insufficient image support.} We observe consistent improvements in
classification and segmentation metrics as we recursively iterate the
core MS-FV-DO architecture.

\subsubsection{Recursive MS-FV-DO Architecture with Local Error Density (LED) Weighting}
 \label{sec:recursive_weighted_classifiers} 

\begin{figure}[t]
  \centering
  \begin{subfigure}{0.13\textwidth}
    \frame{\includegraphics[width=\textwidth]{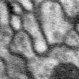}}\vspace{1mm}
    \frame{\includegraphics[width=\textwidth]{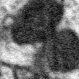}}
    \caption*{image data\\~}
  \end{subfigure}
  \hfill
  \begin{subfigure}{0.13\textwidth}
    \frame{\includegraphics[width=\textwidth]{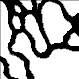}}\vspace{1mm}
    \frame{\includegraphics[width=\textwidth]{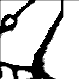}}
    \caption*{ground-truth affinity graph}
  \end{subfigure}
  \hfill
  \begin{subfigure}{0.13\textwidth}
    \frame{\includegraphics[width=\textwidth]{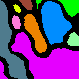}}\vspace{1mm}
    \frame{\includegraphics[width=\textwidth]{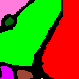}}
    \caption*{ground-truth segmentation}
  \end{subfigure}
  \hfill
  \begin{subfigure}{0.13\textwidth}
    \frame{\includegraphics[width=\textwidth]{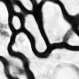}}\vspace{1mm}
    \frame{\includegraphics[width=\textwidth]{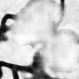}}
    \caption*{MS-FV-DO\\~iter 1}
  \end{subfigure}
  \hfill
  \begin{subfigure}{0.13\textwidth}
    \frame{\includegraphics[width=\textwidth]{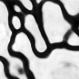}}\vspace{1mm}
    \frame{\includegraphics[width=\textwidth]{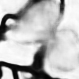}}
    \caption*{MS-FV-DO\\iter 2}
  \end{subfigure}
  \hfill
  \begin{subfigure}{0.13\textwidth}
    \frame{\includegraphics[width=\textwidth]{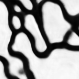}}\vspace{1mm}
    \frame{\includegraphics[width=\textwidth]{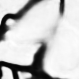}}
    \caption*{MS-FV-DO\\iter 3}
  \end{subfigure}
  \hfill
  \begin{subfigure}{0.13\textwidth}
    \frame{\includegraphics[width=\textwidth]{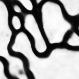}}\vspace{1mm}
    \frame{\includegraphics[width=\textwidth]{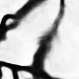}}
    \caption*{MS-FV-DO \emph{w}\\iter 3}
  \end{subfigure}
  \caption{\small Example image data, ground truth, and DAWMR network output for two different locations from the test set. The second row depicts a difficult situation in which mitochondria from distinct cells are directly adjacent; a recursive DAWMR network trained with LED weighting (MS-FV-DO \emph{w} iter 3) is able to interpret the data correctly. }
  \label{fig:affinity_graph_examples}
\end{figure}

Visual inspection of recursive output confirmed that boundary prediction generally improves over multiple iterations, but also revealed that predictions at certain rare image locations did \emph{not} improve. These locations were characterized by a specific property: a high local density of boundary prediction errors present even in the first iteration output. Locally correlated boundary prediction errors are prone to causing mistakes in segmentation and are thus important to avoid. Yet because these locations are rare, they have a negligible impact on boundary prediction accuracy (the metric actually being optimized during training).
Previous work has addressed this issue by proposing learning algorithms that directly optimize segmentation performance~\cite{Turaga:2009, Jain:2010}. These algorithms are computationally expensive, and can make convergence of online gradient descent sensitive to various parameter choices in the loss function and optimization procedure. Therefore we sought a simpler alternative.

\begin{table}
	[htbp] \centering 
	\begin{tabular}
		{|cc|cc|cccc|} \hline & & \multicolumn{2}{c|}{training set (large)} & \multicolumn{4}{c|}{validation set: 16M examples} \\
		model & iter & bal-acc & AUC-edge & bal-acc & AUC-edge & AUC-RI & max RI \\
		\hline MS-FV-DO \emph{w} & 1 & 0.9336 & 0.9818 & 0.8909 & 0.9502 & 0.7178 & 0.9579 \\
		MS-FV-DO \emph{w} & 2 & 0.9453 & 0.9867 & 0.8941 & 0.9524 & 0.7330 & 0.9834 \\
		MS-FV-DO \emph{w} & 3 & 0.9536 & 0.9904 & \textbf{0.9018} & \textbf{0.9606} & \textbf{0.7487} & \textbf{0.9860} \\
		\hline 
	\end{tabular}
	\caption{\small Performance of multiscale foveated architecture over  recursive iterations with LED weighting (\emph{w}).} \label{tbl:recursive_weighting} 
\end{table}

Prior to each recursive iteration we train a DAWMR classifier for 20\%
of the normal number of updates and compute affinity graph output on
the training set. We then create a binary weighting mask with non-zero
entries for each pixel location in which more than 50\% of the
affinity edge classifications in a $5^3$ neighborhood are
incorrect. This simple criteria proves effective in selectively
identifying those rare locations where the failure mode occurred. The
weighting mask is used during training of the full classifier by
sampling weighted locations at a 10x higher rate than normal, and the
mask is combined across iterations by `or'ing. Table
\ref{tbl:recursive_noweighting} shows results from training a
recursive MS-FV-DO architecture with this LED weighting.  Weighting
increased segmentation accuracy, particularly in the second and third
recursive iterations. Boundary prediction classification accuracy was
unaffected or even slightly diminished compared to non-weighted
results. This is consistent with the idea that weighting alters the
cost function to put greater emphasis on specific locations that
influence segmentation accuracy, at the expense of overall boundary
prediction performance.

\subsection{Test Set Evaluation and Comparison} 

\begin{figure}[t]
  \centering
  \includegraphics[width=0.4\textwidth]{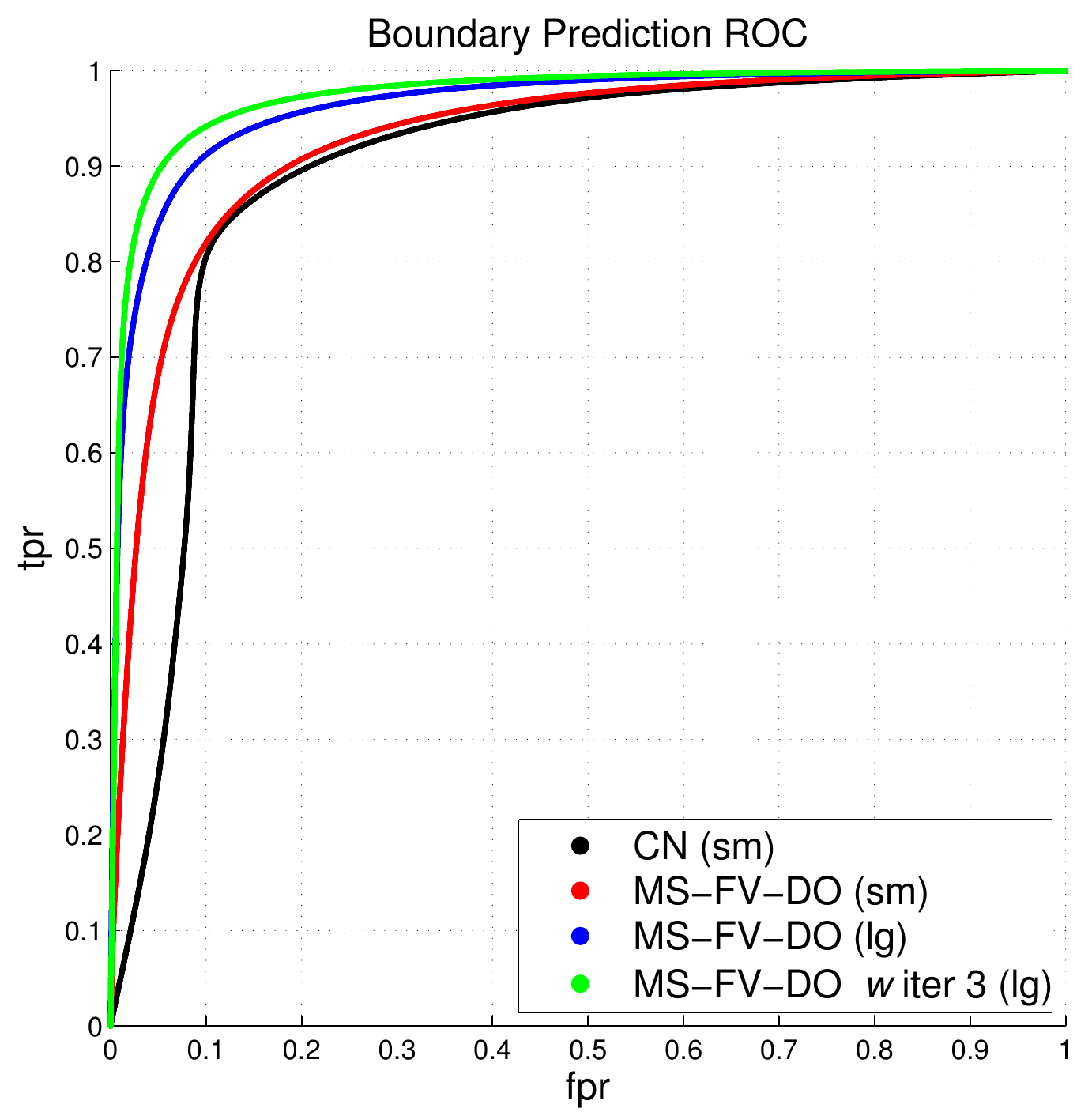}
  \hfill
  \includegraphics[width=0.4\textwidth]{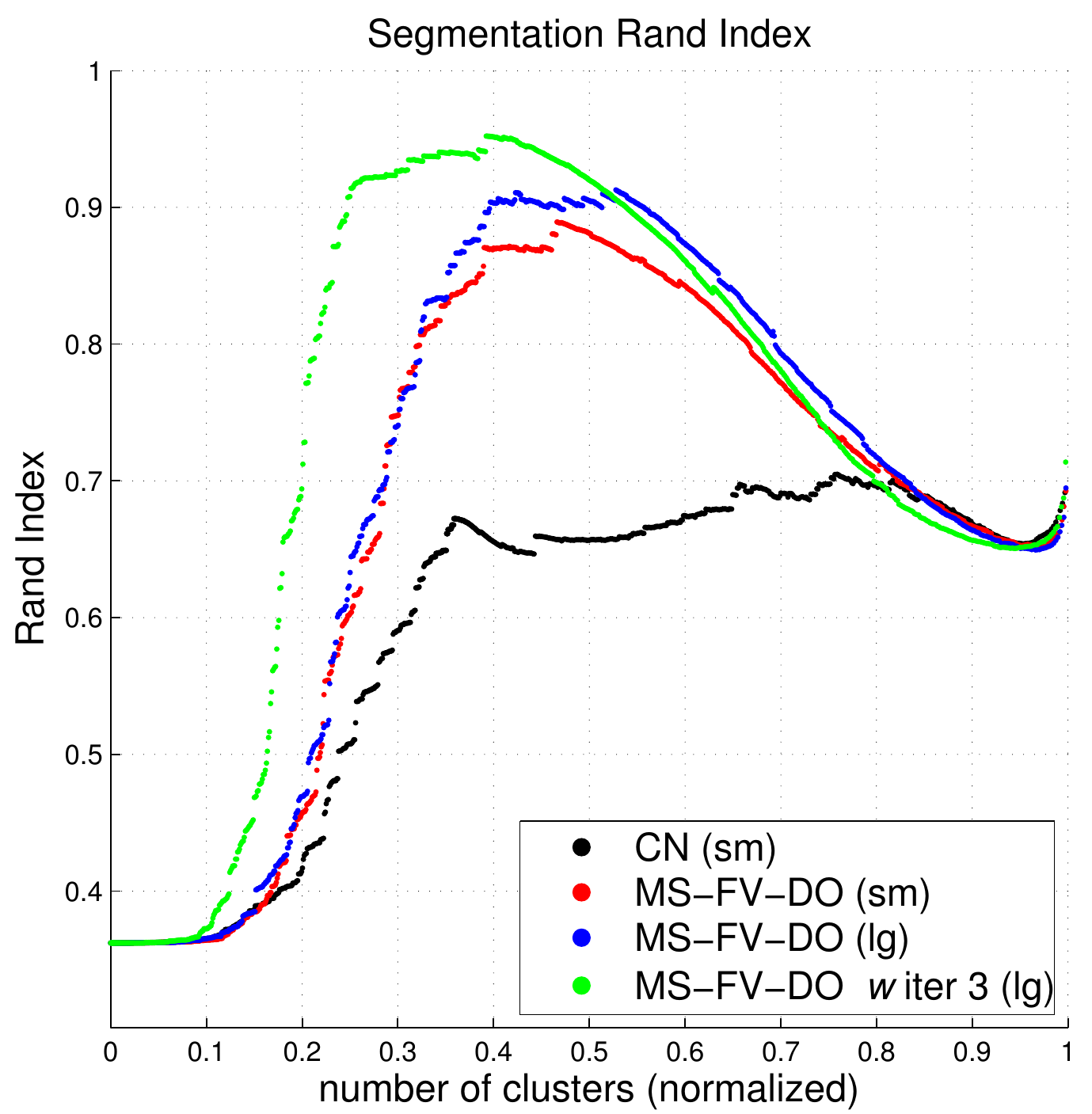}
  \caption{\small Results on the test set. For Rand Index, $x$-axis is
    scaled by the number of clusters in segmentations generated at
    various binarization thresholds (the specific thresholds are
    chosen custom to each methods output, in order to yield
    one-thousand evenly spaced quantiles from the analog affinity
    graph values).}
  \label{fig:test_boundaryroc}
\end{figure}

Based on the experiments performed on the validation set,
we selected a few architectures for evaluation on the full test set.
Details of the architectures are reviewed in Table
\ref{tbl:model_comparison}, summary results are shown in Table
\ref{tbl:final_comparison}, full plots of boundary prediction and
segmentation performance are shown in
Figure~\ref{fig:test_boundaryroc}, and example 2d slices of the
predicted affinity graphs are shown in
Figure~\ref{fig:affinity_graph_examples}.

\begin{table}
	[htbp] \centering 
	\begin{tabular}
		{|l|cccc|} \hline model & $|$unsup$|$ & $|$sup$|$ & fov & training time \\
		\hline CN & 0 & 136000 & $25^3$ & 2 weeks \\
		MS-FV-DO & 250000 & 1600803 & $18^3$ & 1.5 days \\
		MS-FV-DO \emph{w} iter 3 & 750000 & 4802409 & $54^3$ & 5 days \\
		\hline 
	\end{tabular}
	\caption{\small Model comparison of architectures chosen from validation set experiments to be evaluated on the test set. $|$unsup$|$ = number of unsupervised parameters, $|$sup$|$ = number of supervised parameters, and fov = field of view used to generate boundary predictions for a single image location.} \label{tbl:model_comparison} 
\end{table}

The test set results are consistent with the validation set
experiments. Using a 5-million example training set (`sm'), the
MS-FV-DO architecture outperforms the CN with far less training
time. Switching to a larger training set (`lg') improves MS-FV-DO
boundary prediction performance.\footnote{Technical limitations in our
  GPU implementation of convolutional networks prevented us from being
  able to train the CN with the large (lg) training set.} Recursive
iterations and LED weighting further improves segmentation performance
of the DAWMR architecture quite substantially.

\begin{table}
	[htbp] \centering 
	\begin{tabular}
		{|l|ccc|cccc|} \hline & \multicolumn{3}{c|}{training} & \multicolumn{4}{c|}{test set: 46M examples} \\
		model & tr set & bal-acc & AUC-edge & bal-acc & AUC-edge & AUC-RI & max RI \\
		\hline CN & sm & 0.9555 & 0.9872 & 0.8388 & 0.8943 & 0.5927 & 0.7048 \\
		MS-FV-DO & sm & 0.9497 & 0.9899 & 0.8445 & 0.9293 & 0.6798 & 0.8894 \\
		\hline MS-FV-DO & lg & 0.9292 & 0.9809 & 0.9012 & 0.9627 & 0.6939 & 0.9128 \\
		MS-FV-DO \emph{w} iter 3 & lg & 0.9536 & 0.9904 & \textbf{0.9226} & \textbf{0.9735} & \textbf{0.7383} & \textbf{0.9522}\\
		\hline 
	\end{tabular}
	\caption{\small Performance on the full test set for CN and DAWMR architectures.} \label{tbl:final_comparison} 
\end{table}

The results also confirm previous observations that small differences
in boundary prediction accuracy may be associated with large
differences in segmentation accuracy~\cite{Jain:2010}. For example,
the non-recursive MS-FV-DO architecture outperforms the convolutional
network only slightly when measured by \emph{AUC-edge}, but much more
substantially under Rand Index metrics. Visual inspection revealed
that the convolutional network affinity graphs are more prone to
generating undersegmentation errors due to false positive affinity
edges between distinct objects.

\section{Discussion}

{\bf Diverse strategies for exploiting image context:} The DAWMR
networks explored in this work use several different strategies for
manipulating the size of the field of view: multiscale processing,
pooling, and recursive iteration. It is likely that each strategy
offers different modeling capabilities and benefits. For example,
multiscale processing is an efficient way to model image features that
appear at fundamentally different scales, while recursive processing
of the image and affinity graph may be more effective for careful
integration of high-frequency features (e.g., contour completion).

In the supplementary, Table \ref{tbl:larger_unsupervised_fov} shows
that a non-recursive architecture that achieves very large field of
view in a single iteration performs
worse compared to output of a third iteration recursive architecture
with a smaller total field of view. As we lack an
overall theory for the design of such networks, finding the
architecture that makes optimal use of image context requires
empirical model selection.

{\bf A spectrum of weak vs fine tuning in feature learning schemes:}
We employ simple unsupervised learning algorithms to learn features in
DAWMR networks. These features are likely to be only `weakly' tuned
for a specific prediction task, as compared to the `finely' tuned
features learned in a convolutional network trained by supervised
backpropogation. The trade-off, which our empirical results suggest
are well worth it for the problem of boundary prediction, is in the
size of the representation -- DAWMR networks can quickly learn
thousands of features, whereas for convolutional networks it is
currently only practical to use a few dozen at most. Improvements in
computing hardware, or fundamentally more parallel versions of
stochastic gradient descent may enable larger convolutional network
architectures in the future~\cite{niu2011hogwild,le2011building}.

{\bf Recursive iterations and end-to-end learning:} In recursive DAWMR
networks, each iteration is learned without regard to
future iterations. This is in contrast to true `end-to-end' learning,
in which each step
is optimized based on updates back-propagated from the final
output~\cite{LeCun:1998,muller2005off}. While end-to-end learning may
lead to superior discriminative performance, the cost is twofold: a
requirement for using processing stages that are (at least
approximately) differentiable, as well as the computational expense of
performing a `forward' and `backward' pass through all steps for each
parameter update.

We avoid end-to-end learning primarily to minimize training time, but
the freedom to use non-differentiable processing steps in conjunction
with intermediate affinity graphs presents interesting
opportunities. For example, affinity graphs from intermediate
iterations could be converted into segmentations from which
object-level geometric and morphological features are computed. These
features, which may be difficult to represent via differentiable
operations such as filtering (e.g., geodesic and histogram-based
measures), could be used as additional input to further recursive
iterations that refine the affinity graph. This strategy for
exploiting object-level representations is an alternative to
superpixel-based approaches, and may more easily enable correction of
labeling errors that lead to undersegmentation, which is difficult to
address in superpixel approaches.

{\small
\textbf{Acknowledgements:} We thank Zhiyuan Lu for sample preparation, Shan Xu and Harald Hess for FIB-SEM imaging, and Corey Fisher and Chris Ordish for data annotation.
}

{\small{\bibliographystyle{ieee}
\bibliography{curr_opinion}
}}

\newpage

\appendix

\section{Supplementary: Additional Model Selection}

We report the results of additional model selection experiments for
architectures and learning parameters that were less central to
achieving the highest performance in the main presentation.

\subsection{Effect of Training Set Size} 
\label{sec:training_set_size_aug}

As noted in Section~\ref{sec:training_test_evaluation}, two training sets were produced, a small training set (5.2M examples) and a full training set (120M examples), in order to examine how training set size affects DAWMR classification performance.

We augment the full training set by transforming the original data to create synthetic training examples. Specifically, we apply rotations and reflections to the original image data and labels, using a total of seven additional transformations to augment the full training set by a factor of eight. Given the large size of the augmented full training set ($120M*8$), we also subsample examples within each densely labeled subvolume in order to reduce computational load. Training examples that are nearby spatially are likely to have similar statistics, and thus we found that we can achieve comparable performance while using only a subset (10\%) of the full training data.\footnote{The training data consists of many densely labeled subvolumes; using only 10\% of the subvolumes would lead to a much different and less informative training set as compared to the subsampling scheme we propose -- using \emph{all} labeled subvolumes and randomly sampling 10\% of the examples within each.} 

The augmented, subsampled version of the full training data constitutes the `large' (lg) training set referenced in further experiments. Performance while varying the training set is shown in Table~\ref{tbl:single_iter_training}. 
\begin{table}
	[htbp] \centering 
	\begin{tabular}
		{|l|ccc|cccc|} \hline & \multicolumn{3}{c|}{training} & \multicolumn{4}{c|}{validation set: 16M examples} \\
		model & tr set & bal-acc & AUC-edge & bal-acc & AUC-edge & AUC-RI & max RI \\
		\hline SS-FV-DO & sm & 0.9139 & 0.9737 & 0.8383 & 0.9175 & 0.6989 & 0.9411 \\
		SS-FV-DO & lg & 0.9092 & 0.9721 & 0.8667 & 0.9386 & 0.6907 & 0.9311 \\
		\hline MS-FV-DO & sm & 0.9497 & 0.9899 & 0.8372 & 0.9196 & 0.7119 & 0.9327 \\
		MS-FV-DO & lg & 0.9292 & 0.9809 & \textbf{0.8833} & \textbf{0.9497} & \textbf{0.7171} & \textbf{0.9565} \\
		\hline 
	\end{tabular}
	
	\caption{\small Performance of single iteration DAWMR classifiers when varying the size of the training set.} \label{tbl:single_iter_training} 
\end{table}

Expanding the training set results in a significant increase in boundary prediction classification accuracy, but has a somewhat ambiguous impact on segmentation performance. These results suggest that, as training sets become  large, improvements in segmentation accuracy may require additional model capacity or learning algorithms more explicitly focused on segmentation performance. We investigate both of these issues in subsequent experiments.

\subsection{Deeper Feature Extraction Stage}
\label{sec:deeper_feature_extraction}

In the DAWMR architectures discussed in the main text, the
unsupervised stage had a layer of filtering, followed by encoding and
pooling. We experimented with adding a second set of filtering,
encoding, and pooling steps. This modification adds the ability to
learn higher-order image features from the data, and also dramatically
increases the field of view of a single iteration architecture. We
used a pairwise-similarity scheme to group first layer
filters~\cite{coates2011selecting}.

\begin{table}
	[htbp] \centering 
\begin{tabular}{|lc|cc|cccc|} \hline
 & & \multicolumn{2}{c|}{training set (large)} & \multicolumn{4}{c|}{40\% of validation set } \\
model &  fov & bal-acc & AUC-edge & bal-acc & AUC-edge & AUC-RI & max RI \\
\hline
MS-FV-DO     & $18^3$ & 0.9292 & 0.9809 & 0.8831 & 0.9496 & 0.7186 & 0.9464 \\
MS-FV-DO iter 3 & $54^3$ & 0.9473 & 0.9883 & 0.9021 & 0.9628 & 0.7371 & 0.9633 \\
\hline 
MS-FV-DO-DFE     & $62^3$ & 0.9373 & 0.9851 & 0.8876 & 0.9539 & 0.7263 & 0.9627 \\
\hline
\end{tabular}
	\caption{\small Performance of DAWMR architectures from the
          main text compared to an architecture with a deeper feature
          extraction stage (MS-FV-DO-DFE). Validation numbers are on a
          subset of the validation set and thus \emph{not} comparable
          to numbers in the main text (a subset was used due to
          clipping effects caused by the large field of view of the
          MS-FV-DO-DFE
          architecture).} \label{tbl:larger_unsupervised_fov}
\end{table}

We find that this deeper single-iteration architecture, MS-FV-DO-DFE,
improves performance over the standard architecture
(MS-FV-DO). However, performance of the recursive architecture is
superior, even while using less image context, suggesting that
immediately jumping to a large field of view based on deeper
unsupervised feature extraction is not necessarily ideal. We also note
that inference in the MS-FV-DO-DFE architecture is significantly more
computational expensive than the MS-FV-DO architecture, due to the
much larger number of filtering computations in the feature extraction
stage.

\subsection{Varying Feature Dimensionality}

We experimented with varying the dimensionality of the feature
representation produced by the unsupervised feature extraction stage
of the DAWMR networks, by varying the size of the dictionary used for
vector quantization.  The results, shown in
Table~\ref{tbl:varying_feature_dim}, confirm our general hypothesis
that wider networks produced by using a large dictionary yield
increased performance.

\begin{table}[htbp]
  \centering
  \begin{tabular}{|lr|cc|cccc|} \hline
 & & \multicolumn{2}{c|}{training set (large)} & \multicolumn{4}{c|}{validation set: 16M examples} \\
model & dim. & bal-acc & AUC-edge & bal-acc & AUC-edge & AUC-RI & max RI \\
\hline
MS-FV-DO &   400 & 0.9103 & 0.9727 & 0.8710 & 0.9405 & 0.6995 & 0.9236 \\
MS-FV-DO &   800 & 0.9176 & 0.9749 & 0.8760 & 0.9466 & 0.7078 & 0.9444 \\
MS-FV-DO &  2000 & 0.9247 & 0.9775 & 0.8810 & 0.9467 & 0.7027 & 0.9533 \\
MS-FV-DO &  4000 & 0.9270 & 0.9793 & 0.8828 & 0.9468 & 0.7136 & 0.9452 \\
MS-FV-DO &  8000 & 0.9292 & 0.9809 & 0.8833 & 0.9497 & 0.7171 & 0.9565 \\ 
MS-FV-DO & 12000 & 0.9301 & 0.9803 & 0.8878 & 0.9495 & 0.7144 & 0.9375 \\
\hline
  \end{tabular}
  \caption{Performance of DAWMR architectures when varying the
    dimensionality of the feature representation from the unsupervised
    stage.}
  \label{tbl:varying_feature_dim}
\end{table}

\subsection{Varying the Number of MLP Hidden Units}

We also experimented with varying the number of hidden units used in
the supervised MLP classifier, with results shown in
Table~\ref{tbl:varying_mlp_hu}.  All classifiers were trained using
the same fixed number of updates.  In general we found the results to
not be especially sensitive to this parameter, and used 200 as a
balance between sufficient capacity and faster training and
convergence.

\begin{table}[htbp]
  \centering
  \begin{tabular}{|lc|cc|cccc|} \hline
 & & \multicolumn{2}{c|}{training set (large)} & \multicolumn{4}{c|}{validation set: 16M examples} \\
model & \# h.u. & bal-acc & AUC-edge & bal-acc & AUC-edge & AUC-RI & max RI \\
\hline
MS-FV-DO &  50 & 0.9225 & 0.9770 & 0.8793 & 0.9467 & 0.7197 & 0.9444 \\
MS-FV-DO & 100 & 0.9279 & 0.9795 & 0.8849 & 0.9491 & 0.7063 & 0.9532 \\
MS-FV-DO & 200 & 0.9292 & 0.9809 & 0.8833 & 0.9497 & 0.7171 & 0.9565 \\ 
MS-FV-DO & 400 & 0.9278 & 0.9762 & 0.8833 & 0.9438 & 0.7027 & 0.9272 \\
\hline
  \end{tabular}
  \caption{Performance of DAWMR architectures when varying the number
    of hidden units (h.u.) used in supervised MLP.}
  \label{tbl:varying_mlp_hu}
\end{table}

\subsection{Varying the Number of MLP Hidden Layers}

We experimented with adding additional layers of hidden units in the
supervised MLP classifier, with results shown in
Table~\ref{tbl:mlp_multiple_hidden_layers}.  The network was kept at a
fixed with of 200 hidden units at each hidden layer, and a drop-out
rate of 0.5 was used at each hidden layer.  All classifiers were
trained using the same fixed number of updates.

\begin{table}[htbp]
  \centering
  \begin{tabular}{|lc|cc|cccc|} \hline
 & & \multicolumn{2}{c|}{training set (large)} & \multicolumn{4}{c|}{validation set: 16M examples} \\
model & \# h.l. & bal-acc & AUC-edge & bal-acc & AUC-edge & AUC-RI & max RI \\
\hline
MS-FV-DO & 1 & 0.9330 & 0.9820 & 0.8890 & 0.9521 & 0.7206 & 0.9519 \\
MS-FV-DO & 2 & 0.9363 & 0.9848 & 0.8910 & 0.9560 & 0.7308 & 0.9547 \\
MS-FV-DO & 3 & 0.9396 & 0.9851 & 0.8945 & 0.9594 & 0.7393 & 0.9586 \\
MS-FV-DO & 4 & 0.9323 & 0.9829 & 0.8888 & 0.9563 & 0.7311 & 0.9476 \\
\hline
  \end{tabular}
  \caption{Performance of DAWMR architectures when varying the number
    of hidden layers (h.l.) used in supervised MLP.}
  \label{tbl:mlp_multiple_hidden_layers}
\end{table}


\subsection{Whitening}

Previous work with vector quantization and deep learning architectures
has noted the importance of whitening the data prior to dictionary
learning~\cite{coates2011importance}.  We experimented with adding
contrast normalization and ZCA whitening to the DAWMR networks.  As
shown in Table~\ref{tbl:adding_whcn}, we generally found that both
contrast normalization and whitening generally decreased performance
slightly.  These results seem to indicate the importance of keeping
information about intensity values relative to the global data rather
than just a local patch, for this particular data set and in
distinction to other data such as natural images.

For DAWMR networks with multiple feature encoding steps, as presented
in Section~\ref{sec:deeper_feature_extraction}, we have had success
with combining a small number of features produced by a single VQ step
and no whitening with a larger number of features produced by multiple
VQ steps and whitening.

\begin{table}[htbp]
  \centering
  \begin{tabular}{|l|cc|cccc|} \hline
 & \multicolumn{2}{c|}{training set (large)} & \multicolumn{4}{c|}{validation set: 16M examples} \\
model & bal-acc & AUC-edge & bal-acc & AUC-edge & AUC-RI & max RI \\
\hline
MS-FV-DO + CN,WH & 0.9191 & 0.9746 & 0.8722 & 0.9411 & 0.7039 & 0.9142 \\
MS-FV-DO         & 0.9292 & 0.9809 & 0.8833 & 0.9497 & 0.7171 & 0.9565 \\ 
\hline
  \end{tabular}
  \caption{Performance of DAWMR architecture when adding contrast
    normalization (CN) and ZCA whitening (WH).}
  \label{tbl:adding_whcn}
\end{table}

\subsection{Orthogonal Matching Pursuit vs K-means}

In initial experiments, we used K-means for dictionary learning and
``triangle K-means'' for feature encoding~\cite{coates2011importance}.
This is compared with the Orthogonal Matching Pursuit that we used in
the main presentation in Table~\ref{tbl:kmeans_omp}.  In general, we
found both to give comparable results, with OMP allowing for faster
feature encoding and seeming to be more amenable to multiple layers of
feature encoding (Section~\ref{sec:deeper_feature_extraction}).

\begin{table}[htbp]
  \centering
  \begin{tabular}{|lc|cc|cccc|} \hline
 & & \multicolumn{2}{c|}{training set (large)} & \multicolumn{4}{c|}{validation set: 16M examples} \\
model & learning & bal-acc & AUC-edge & bal-acc & AUC-edge & AUC-RI & max RI \\
\hline
MS-FV-DO & K-means & 0.9317 & 0.9819 & 0.8852 & 0.9499 & 0.7195 & 0.9321 \\
MS-FV-DO & OMP     & 0.9292 & 0.9809 & 0.8833 & 0.9497 & 0.7171 & 0.9565 \\ 
\hline
  \end{tabular}
  \caption{Performance of DAWMR architectures when varying the
    unsupervised stage dictionary learning method and feature
    encoding.}
  \label{tbl:kmeans_omp}
\end{table}

\section{Supplementary: Network Details}

Here we present details and values of parameters used in our models.

\subsection{Convolutional Network}
The convolutional network was trained in accordance with procedures
outlined in previous work~\cite{Jain:2007, Turaga:2010uq}. We used
sigmoid units and performed greedy layer-wise training of the
architecture: 5e5 updates after adding each layer, 2e6 updates for the
final architecture. Networks trained with significantly fewer
iterations exhibited much worse training set performance. During
training, we used a balanced sampling strategy that alternated between
negative and positive edge locations and selected a $5^3$ cube around
each edge as a minibatch. Learning rates were set to $0.1$, except for
the last layer (set to $0.01$). A square-square
loss~\cite{Turaga:2010uq, Jain:2010} was optimized with a margin of
$0.3$.

\subsection{Multilayer Perceptron}

The multilayer perceptrons in DAWMR architectures were trained using
minibatch sizes of $40$ with a balanced sampling of positive and
negative edges. Learning rates were set to $0.02$. We used sigmoid
output units and rectified linear units in the hidden layer. For
networks trained with drop-out regularization, the drop-out rate was
set to $0.5$ for the hidden layer and $0$ for the input layer. We
performed 5e5 updates. Optimization was performing using a
cross-entropy loss function.  To regularize and prevent overfitting,
we used an ``inverse margin'' of 0.1, meaning that target labels were
set to $0.1/0.9$ rather than $0/1$, penalizing over-confident
predictions.



\section{Supplementary: DAWMR Implementation and Training Time}

In this section we describe the code implementation details and
training time analysis for DAWMR networks.

\subsection{Implementation}
\label{sec:implementation}

\begin{figure}
	[t] \centering 
	\includegraphics[width= 
	\textwidth]{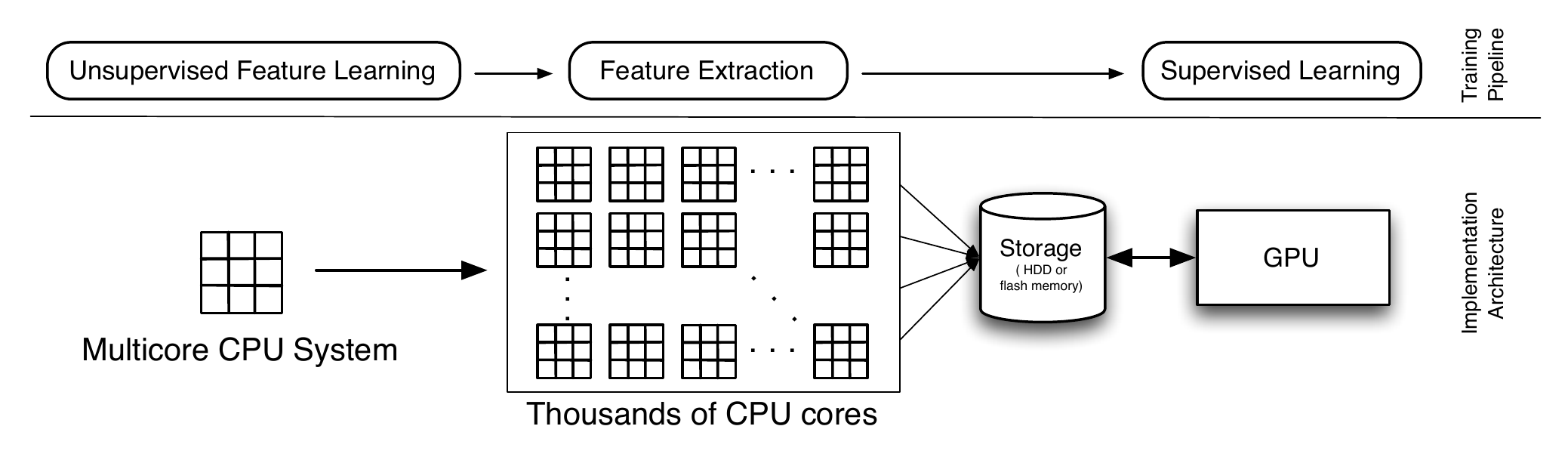} \caption{\small Computation architecture used in experiments. A CPU cluster is used to pre-compute feature vectors prior to GPU-based training of a supervised classifier.} \label{fig:dawmr_computational_architecture} 
\end{figure}

The design of DAWMR networks permits the use of parallel computing strategies that result in fast training time. A schematic illustration of our pipeline is given in Figure~\ref{fig:dawmr_computational_architecture}. 

Unsupervised feature learning is performed on a traditional multicore CPU. Next, features are extracted for potentially millions of locations  distributed across a large 3d image volume with billions to trillions of voxels. In our experiments, this computation is spread across a CPU cluster comprising thousands of cores. Each worker loads image data for the locations it has been assigned, extracts features, and writes the final feature vector to a file or distributed database. Lastly, supervised learning is performed on a single GPU-equipped machine by repeatedly loading a random selection of feature vectors and performing online minibatch gradient updates with a GPU implementation of a multilayer perceptron.

Gradient-based supervised training of deep convolutional networks requires performing forward and backward pass computations through many layers of processing, and the intricate nature of these computations limits the extent of parallelism that can typically be achieved. By learning the feature representation via efficient unsupervised algorithms, DAWMR networks are able to `pre-compute' feature vectors for each example in parallel across a large CPU cluster (in our experiments, this phase of computation can be accomplished in tens of minutes for even one-hundred million examples).
 
An open source Matlab/C software package that implements DAWMR networks is available online: \url{http://sites.google.com/site/dawmrlib/}. 

\subsection{Training Time}

Training DAWMR networks with parallel computation hardware (a CPU cluster and GPUs) results in training times on the order of a single day for a single iteration classifier, and multiple days for multiple recursive iterations. This compares favorably with purely supervised multilayer convolutional networks (typically on the order of weeks for GPU implementations with 3d filters, even without multiscale processing). 

An analysis of our pipeline (Figure~\ref{fig:dawmr_computational_architecture})  reveals that the vast majority of time is spent training the multilayer perceptron (MLP). Moreover, during the GPU-based MLP training, most of the time is spent on I/O to retrieve feature vectors from the filesystem for each randomly constructed minibatch. In our experiments, the filesystem was a large-scale EMC Isilon installation accessed via 10-gigabit networking.

Substantial improvements in training time could thus be achieved by additional engineering that simply reduced the overhead associated with accessing feature vectors. In-memory databases, flash storage, and more efficient distributed filesystems are likely to enable such improvements.

\end{document}